\documentclass{article}
\usepackage{spconf,amsmath,graphicx,hyperref}
\usepackage{enumitem, bm, amssymb, booktabs, multirow, cite, setspace, etoolbox, xcolor, marvosym}

\newcommand{\red}[1]{\textcolor{red}{#1}}
\newcommand{\blue}[1]{\textcolor{blue}{#1}}

\title{ACD-CLIP: Decoupling Representation and Dynamic Fusion for Zero-Shot Anomaly Detection}
\name{
	Ke Ma\textsuperscript{1}\quad
	Jun Long\textsuperscript{1}\quad
	Hongxiao Fei\textsuperscript{1}\quad
	Liujie Hua\textsuperscript{1}\quad
	Zhen Dai\textsuperscript{2}\quad
	Yueyi Luo\textsuperscript{1*}\thanks{* Corresponding author. E-mail: luoyueyi@csu.edu.cn}
}

\address{
	$^{1}$Central South University, Changsha, China\\
	$^{2}$Hunan Vocational College of Science and Technology, Changsha, China
}

\makeatletter
\patchcmd{\@maketitle}{\vskip 1.5em}{\vskip 1em}{}{}
\patchcmd{\@maketitle}{\@name \\ \@address}{\@name \\[-1em] \@address}{}{}
\patchcmd{\@maketitle}{\vskip 1.5em}{\vskip 0.2em}{}{}
\makeatother
\begin{document}
	\maketitle
	\setlength{\abovedisplayskip}{5pt}
	\setlength{\belowdisplayskip}{5pt}
	\begin{abstract}
		\vspace{-0.5em}
		Pre-trained Vision-Language Models (VLMs) struggle with Zero-Shot Anomaly Detection (ZSAD) due to a critical adaptation gap: they lack the local inductive biases required for dense prediction and employ inflexible feature fusion paradigms. We address these limitations through an \textbf{A}rchitectural \textbf{C}o-\textbf{D}esign framework that jointly refines feature representation and cross-modal fusion. Our method proposes a parameter-efficient \textbf{Convolutional Low-Rank Adaptation (Conv-LoRA)} adapter to inject local inductive biases for fine-grained representation, and introduces a \textbf{Dynamic Fusion Gateway (DFG)} that leverages visual context to adaptively modulate text prompts, enabling a powerful bidirectional fusion. Extensive experiments on diverse industrial and medical benchmarks demonstrate superior accuracy and robustness, validating that this synergistic co-design is critical for robustly adapting foundation models to dense perception tasks. The source code is available at~\url{https://github.com/cockmake/ACD-CLIP}.
	\end{abstract}

	\begin{figure}[!t]
		\centering
		\includegraphics[width=0.85\linewidth]{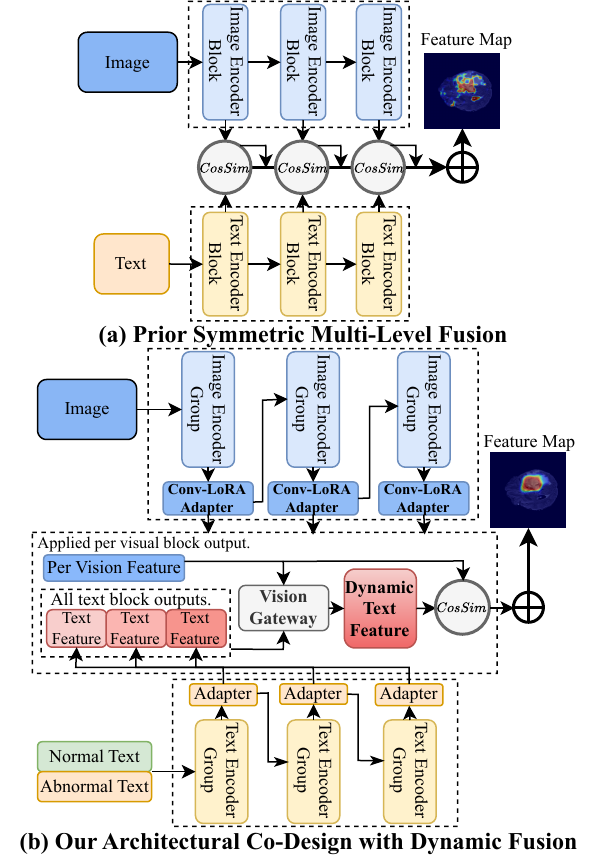}
		\vspace{-1.5em}
		\caption{
			\textbf{Comparison of fusion paradigms.} 
			(a) Prior works rely on a rigid, static alignment between corresponding feature blocks. 
			(b) Our Architectural Co-Design enables a flexible fusion policy by enriching visual features with local priors (Conv-LoRA) and then dynamically generating tailored text descriptors for each visual level (DFG).
		}
		\vspace{-1.5em}
		\label{fig:1}
	\end{figure}
	
	\begin{figure*}[!t]
		\centering
		\includegraphics[width=0.95\textwidth]{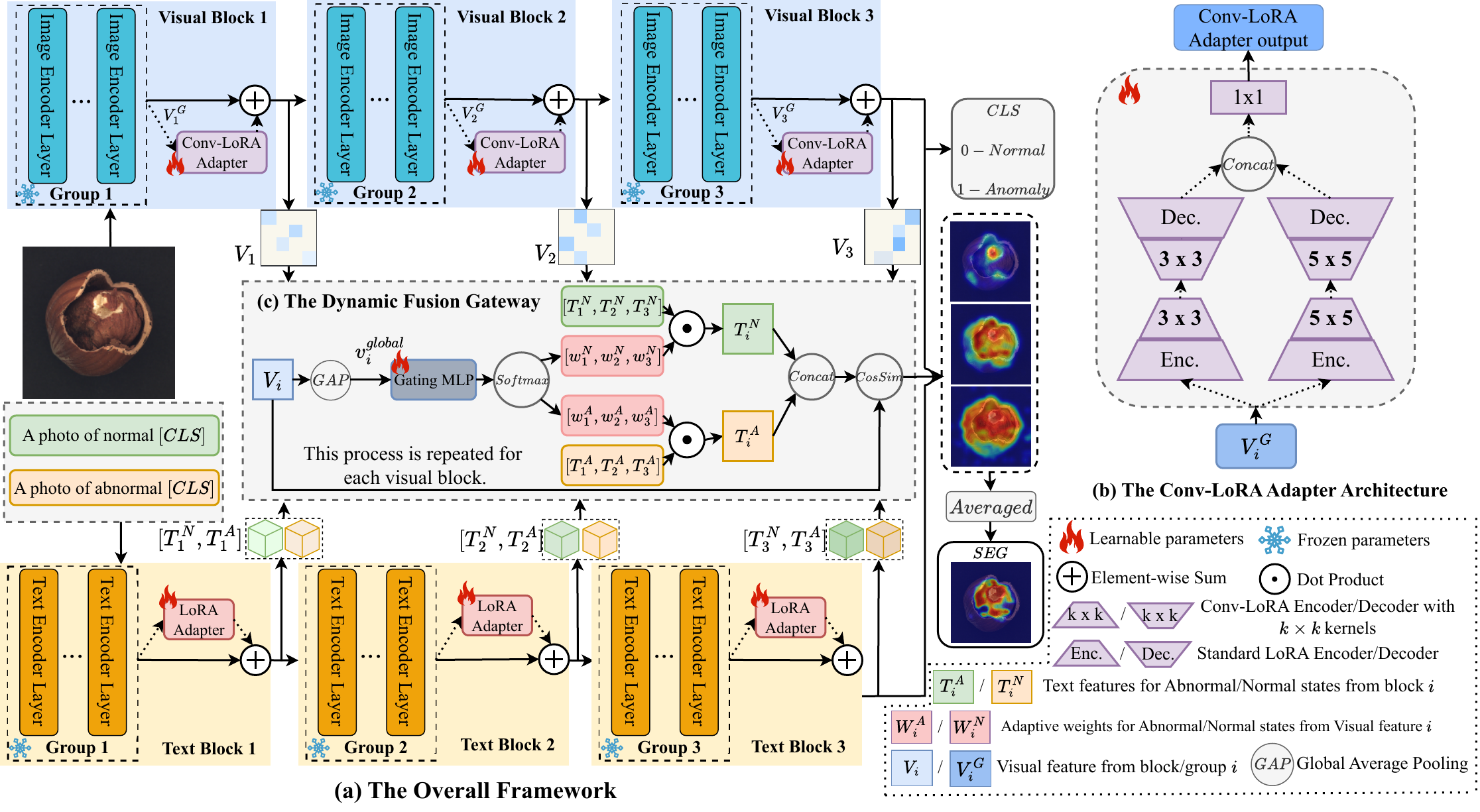}
		\vspace{-1.5em}
		\caption{
			\textbf{Overview of the proposed ACD-CLIP architecture.}
			\textbf{The Overall Framework:} We structure CLIP's vision and text encoders into a hierarchy of $N$ sequential \textbf{Groups} (as illustrated, $N = 3$). Each vision group is enhanced by a trainable \textbf{Conv-LoRA Adapter} to instill local priors, while each corresponding text group incorporates a standard LoRA adapter. \textbf{The Dynamic Fusion Gateway} then uses each visual feature $V_i$ to generate a tailored text descriptor for producing a level-specific anomaly map.
			\textbf{The Conv-LoRA Adapter:} Our parameter-efficient adapter features a multi-branch design with multi-scale convolutions inside a LoRA bottleneck.
		}
		\vspace{-1.5em}
		\label{fig:2}
	\end{figure*}
	
	\vspace{-0.9em}
	\section{Introduction}
	\label{sec:intro}
	\vspace{-0.65em}
	Zero-Shot Anomaly Detection (ZSAD) adapts Vision-Language Models (VLMs) \cite{flamingo, liu2023visual} like CLIP \cite{clip} to circumvent the extensive training data required by traditional methods \cite{roth2022towards, defard2021padim}. The dominant paradigm uses text prompts, evolving from hand-crafted ensembles to learnable prompts tuned on auxiliary data (WinCLIP \cite{winclip}). These advanced methods have explored object-agnostic semantics (AnomalyCLIP \cite{zhou2023anomalyclip}), dynamic prompt generation (AdaCLIP \cite{adaclip}), and multi-layer feature queries (CLIP-AD \cite{clipad}). However, this line of work faces two fundamental limitations.
	
	First, these methods rely on an \textbf{inflexible fusion paradigm} [Figure~\ref{fig:1}(a)], treating the VLM as a black box. This static, layer-wise alignment assumes a fixed semantic correspondence, which restricts transferability by enforcing rigid alignments that fail when anomalies shift semantic scales across different domains. Second, a deeper \textbf{representational adaptation gap} exists, as CLIP's ViT \cite{vit} architecture lacks the local inductive biases inherent in CNNs \cite{long2015fully, ronneberger2015u}. This gap, stemming from CLIP’s focus on global semantics, critically limits anomaly discriminability for fine-grained structural changes, leading to the loss of subtle local features essential for precise dense prediction.
	
	To address these core issues, we introduce the \textbf{A}rchitectural \textbf{C}o-\textbf{D}esign (ACD-CLIP) framework [Figure~\ref{fig:1}(b)], a novel approach that synergistically refines both feature representation and cross-modal fusion. The synergy is crucial: the Conv-LoRA adapter provides fine-grained details that are fully exploited by the DFG's adaptive fusion mechanism. Our architectural modifications are enabled by Parameter-Efficient Fine-Tuning (PEFT) \cite{houlsby2019parameter, lora}. While recent works have shown the value of integrating convolutional structures into PEFT adapters for general vision tasks \cite{jie2024convolutional, convlora}, our work is the first to co-design such an adapter with a dynamic fusion mechanism specifically for ZSAD. Our contributions are: 1) A parameter-efficient \textbf{Conv-LoRA Adapter} to inject local inductive biases for fine-grained representation. 2) A \textbf{Dynamic Fusion Gateway}, a vision-guided mechanism enabling a flexible, bidirectional fusion policy. 3) Validation of our method, which yields \textbf{significant performance gains} on diverse industrial and medical ZSAD benchmarks.

	\vspace{-1.2em}
	\section{Methodology}
	\label{sec:methodology}
	\vspace{-0.8em}
	As illustrated in Figure~\ref{fig:2}, our ACD-CLIP framework introduces two core innovations to adapt VLMs for ZSAD: \textbf{(1)} a \textbf{Conv-LoRA Adapter} to instill local inductive biases for fine-grained representation, and \textbf{(2)} a \textbf{Dynamic Fusion Gateway} for adaptive cross-modal fusion.
	
	\vspace{-1em}
	\subsection{Hierarchical Feature Adaptation with Local Priors}
	\vspace{-0.5em}
	To mitigate feature entanglement, we partition the vision and text encoders into $N$ sequential \textbf{Groups}, encouraging each to specialize in representations at a distinct semantic level. To instill local inductive priors, we integrate a parameter-efficient \textbf{Conv-LoRA Adapter} into each vision group. Distinct from prior work that utilizes a Mixture-of-Experts approach for dynamic feature scaling \cite{convlora}, our adapter employs a parallel multi-branch architecture with distinct kernel sizes, specifically tailored to capture the varied local patterns essential for fine-grained anomaly detection. The adapter processes the sequential output from a ViT block, $X_{\text{in}} \in \mathbb{R}^{B \times L \times C}$, where $B$ is the batch size, $L$ is the sequence length, and $C$ is the channel dimension.
	
	The adapter features a parallel multi-branch design with distinct $k \times k$ convolutional kernels (where $k \in \{3, 5\}$) that operate within a LoRA-style bottleneck. This structure allows it to capture local patterns at varied receptive fields while maintaining parameter efficiency. The entire operation, $\Phi_{\text{Adapter}}$, generates a residual update $\Delta X$ by fusing the outputs of the parallel branches:
	\begin{equation}
		\begin{aligned}
			\Delta X &= \Phi_{\text{Adapter}}(X_{\text{in}}) \\
			&= \mathcal{R}_{\text{seq}}\Big(\operatorname{Conv}_{1 \times 1}\big(\mathcal{R}_{\text{2D}}\big( \operatorname{Concat}\big[X_{\text{branch\_out}}^{(k)}\big] \big)\big)\Big)
		\end{aligned}
	\end{equation}
	where $\operatorname{Concat}[\cdot]$ aggregates the feature maps from each branch along the channel dimension, and a subsequent $1 \times 1$ convolution, $\operatorname{Conv}_{1 \times 1}$, adaptively fuses these multi-scale features into a unified residual update.
	
	The output of each branch, $X_{\text{branch\_out}}^{(k)}$, is generated in a two-step process. First, the input tensor $X_{\text{in}}$ is projected to a low-rank space by $W_{\text{down}}$, reshaped, and passed through a $k \times k$ convolution to instill local context and transform the features, producing an intermediate feature map $X_{\text{conv}}^{(k)}$:
	\begin{equation}
		X_{\text{conv}}^{(k)} = \frac{1}{k}\operatorname{Conv}_{\text{down}}^{(k)}\left(\mathcal{R}_{\text{2D}}(X_{\text{in}} W_{\text{down}})\right)
	\end{equation}
	This feature map is then processed by a second $k \times k$ convolution for further feature refinement, reshaped back into a sequence, and projected up by $W_{\text{up}}$ to yield the final branch output:
	\begin{equation}
		X_{\text{branch\_out}}^{(k)} = \mathcal{R}_{\text{seq}}\left(\frac{1}{k}\operatorname{Conv}_{\text{up}}^{(k)}\left(X_{\text{conv}}^{(k)}\right)\right) W_{\text{up}}
	\end{equation}
	Here, $W_{\text{down}}$ and $W_{\text{up}}$ are the LoRA projection matrices performing channel-wise compression and expansion. $\mathcal{R}_{\text{2D}}$ and $\mathcal{R}_{\text{seq}}$ are operators for reshaping between sequence and spatial formats. $\operatorname{Conv}_{\text{up}}^{(k)}$ and $\operatorname{Conv}_{\text{down}}^{(k)}$ are standard $k \times k$ convolutions that operate on feature channels for transformation and local context infusion while preserving spatial dimensions.
	
	\vspace{-1em}
	\subsection{Dynamic Fusion Gateway}
	\label{sec:dynamic_fusion_gateway}
	\vspace{-0.5em}
	To overcome static fusion, our \textbf{Dynamic Fusion Gateway} generates custom semantic descriptors for each visual feature group $V_i$. A global context vector $v_i^{\text{global}} = \operatorname{GAP}(V_i)$ is extracted and passed through a gating MLP to produce logits $\boldsymbol{l}_i^s$ for each semantic state $s \in \{N, A\}$ (Normal, Abnormal). These logits determine the dynamic fusion weights $\omega_{i,j}^s$ for the set of all text features $\{T_j^s\}_{j=1}^N$:
	\begin{equation}
		\omega_{i,j}^s = \frac{\exp(l_{i,j}^s)}{\sum_{k=1}^N \exp(l_{i,k}^s)}, \quad \text{where } \boldsymbol{l}_i^s = \text{MLP}_{\text{gate}}^s(v_i^{\text{global}})
	\end{equation}
	These weights dynamically fuse the multi-level text features to produce context-aware normal ($T_i^N$) and abnormal ($T_i^A$) descriptors:
	\begin{equation}
		T_i^s = \sum_{j=1}^N \omega_{i,j}^s \cdot T_j^s, \quad \text{for } s \in \{N, A\}
	\end{equation}
	The resulting level-specific anomaly map, $M_i$, is then computed by applying a softmax function over the patch-wise cosine similarities between the visual features $V_i$ and their corresponding dynamic text descriptors:
	\begin{equation}
		\label{eq:anomaly_map}
		M_i = \frac{\exp(\text{sim}(V_i, T_i^A) / \tau)}{\exp(\text{sim}(V_i, T_i^N) / \tau) + \exp(\text{sim}(V_i, T_i^A) / \tau)}
	\end{equation}
	where $\text{sim}(\cdot,\cdot)$ denotes the cosine similarity operator and $\tau$ is a temperature parameter. The final anomaly map is the average of these level-specific maps $\{M_i\}$.

	\vspace{-1em}
	\subsection{Training Objective and Inference}
	\vspace{-0.5em}
	The framework is trained end-to-end with a composite loss function, which jointly addresses pixel-level segmentation and image-level classification.
	
	\noindent \textbf{Segmentation Loss.} The final prediction map is the average of level-specific maps, $M_{\text{seg}} = \frac{1}{N} \sum_{i=1}^N M_i$. It is optimized against the ground-truth mask $M_{\text{GT}}$ using a combined Focal~\cite{lin2017focal} and Dice loss~\cite{milletari2016v}:
	\begin{align}
		\mathcal{L}_{\text{seg}} = \lambda_{\text{Focal}} \mathcal{L}_{\text{Focal}}(M_{\text{seg}}, M_{\text{GT}}) \nonumber \\  + \lambda_{\text{Dice}} \mathcal{L}_{\text{Dice}}(M_{\text{seg}}, M_{\text{GT}})
	\end{align}
	where the balancing hyperparameters $\lambda_{\text{Focal}}$ and $\lambda_{\text{Dice}}$ are both set to 1.0.
	
	\noindent\textbf{Classification Loss.} For image-level classification, we compute the cosine similarity between the final global visual feature and the original, unfused text features from the last group, supervised via Cross-Entropy loss $\mathcal{L}_{\text{cls}}$. We use the original text features here to provide a stable, high-level semantic anchor for the classification task.
	
	\noindent\textbf{Total Objective.} The final training objective is a weighted sum of the two losses, with the classification weight $\lambda_{\text{cls}}$ set to 0.5:
	\begin{equation}
		\mathcal{L}_{\text{total}} = \mathcal{L}_{\text{seg}} + \lambda_{\text{cls}} \mathcal{L}_{\text{cls}}
	\end{equation}
	
	\noindent\textbf{Inference.} A single forward pass generates the anomaly map $M_{\text{seg}}$ and a classification score. The final image-level anomaly score combines the direct classification output with the maximum value from $M_{\text{seg}}$.

	\begin{table*}[!t]
		\centering
		\caption{Quantitative comparison with state-of-the-art methods across diverse ZSAD benchmarks. Results are reported as (AUROC, AP) in percentage. \blue{Blue:} Best result. \red{Red:} Second-best result.}
		\label{tab:main_results}
		\footnotesize
		\setlength{\tabcolsep}{5pt} 
		\begin{tabular}{llcccccccc}
			\toprule
			\multirow{2}{*}{\textbf{\begin{tabular}{@{}c@{}}Domain \\ (Metric)\end{tabular}}} & \multirow{2}{*}{\textbf{Dataset}} & \multirow{2}{*}{\textbf{WinCLIP}} & \multirow{2}{*}{\textbf{CLIP-AD}} & \multirow{2}{*}{\textbf{AnomalyCLIP}} & \multirow{2}{*}{\textbf{AdaCLIP}} & \multicolumn{4}{c}{\textbf{ours}} \\
			\cmidrule(lr){7-10} 
			& & & & & & \textbf{$N = 2$} & \textbf{$N = 3$} & \textbf{$N = 4$} & \textbf{$N = 6$} \\
			\midrule
			\multirow{5}{*}{\begin{tabular}{@{}c@{}}Industrial \\ (Pixel-level)\end{tabular}} & MVTec-AD & (85.1, 18.0) & (89.8, 40.0) & (91.1, 34.5) & (86.8, 38.1) & (\blue{91.7}, \blue{44.1}) & (\red{91.4}, 43.6) & (90.9, \red{44.0}) & (90.0, 43.1) \\
			& BTAD & (71.4, 11.2) & (93.1, 46.7) & (93.3, 42.0) & (87.7, 36.6) & (\red{96.3}, \red{51.2}) & (95.9, 51.2) & (\blue{96.5}, \blue{51.5}) & (94.6, 51.1) \\
			& MPDD & (95.2, 28.1) & (95.1, 28.4) & (96.2, 28.9) & (96.6, 29.1) & (\blue{97.0}, 29.8) & (96.3, \red{30.3}) & (96.1, 29.4) & (\red{96.6}, \blue{30.3}) \\
			& RSDD & (95.1, 2.1) & (99.2, 31.9) & (99.1, 19.1) & (\blue{99.5}, 38.2) & (99.1, \red{40.7}) & (\red{99.4}, \blue{41.3}) & (98.9, 40.4) & (98.6, 40.1) \\
			& VisA & (79.6, 5.0) & (95.0, 26.3) & (95.4, 21.3) & (95.1, \red{29.2}) & (\red{95.7}, 27.8) & (\blue{95.9}, \blue{29.6}) & (94.6, 27.5) & (94.1, 27.3) \\
			\cmidrule{1-10}
			\multirow{6}{*}{\begin{tabular}{@{}c@{}}Medical \\ (Pixel-level)\end{tabular}} & ColonDB & (64.8, 14.3) & (80.3, 23.7) & (81.9, 31.3) & (79.3, 26.2) & (\red{85.0}, \blue{35.9}) & (\blue{85.1}, 32.6) & (83.3, \red{35.4}) & (80.7, 31.1) \\
			& ClinicDB & (70.7, 19.4) & (85.8, 39.0) & (85.9, 42.2) & (84.3, 36.0) & (\blue{90.4}, 53.5) & (89.2, \red{54.0}) & (\red{89.8}, \blue{56.1}) & (85.0, 49.1) \\
			& Kvasir & (69.8, 27.5) & (82.5, 46.2) & (81.8, 42.5) & (79.4, 43.8) & (88.5, \red{60.7}) & (\red{88.8}, 60.2) & (\blue{88.8}, \blue{61.3}) & (83.3, 54.8) \\
			& BrainMRI & (86.0, 49.2) & (96.4, 54.2) & (95.6, 53.1) & (93.9, 52.3) & (96.6, 55.6) & (95.3, 53.0) & (\blue{97.0}, \blue{61.0}) & (\red{96.9}, \red{56.1}) \\
			& Liver CT & (96.2, 7.2) & (95.4, 7.1) & (93.9, 5.7) & (94.5, 5.9) & (\blue{97.3}, \blue{8.8}) & (\red{97.2}, \red{7.5}) & (96.0, 6.8) & (95.3, 6.2) \\
			& Retina OCT & (80.6, 39.8) & (90.9, 48.7) & (92.6, \red{55.3}) & (88.5, 47.1) & (\blue{96.1}, \blue{66.2}) & (\red{93.7}, 50.9) & (91.3, 48.2) & (91.5, 48.7) \\
			\midrule
			\multirow{5}{*}{\begin{tabular}{@{}c@{}}Industrial \\ (Image-level)\end{tabular}} & MVTec-AD & (89.3, 92.9) & (89.8, 95.3) & (90.3, 95.1) & (90.7, 95.2) & (\red{90.9}, 95.7) & (90.7, \red{95.8}) & (\blue{92.4}, \blue{96.8}) & (90.7, 95.7) \\
			& BTAD & (83.3, 84.1) & (85.8, 85.2) & (89.1, 91.1) & (91.6, 92.4) & (93.5, 96.0) & (\red{94.9}, \red{98.0}) & (93.3, 94.0) & (\blue{95.4}, \blue{98.2}) \\
			& MPDD & (63.6, 71.2) & (74.5, 77.9) & (73.7, 77.1) & (72.1, 76.9) & (\blue{78.1}, \blue{83.7}) & (\red{77.6}, \red{82.3}) & (74.7, 79.0) & (74.8, 78.2) \\
			& RSDD & (85.3, 65.3) & (88.3, 73.9) & (73.5, 55.0) & (89.1, 70.8) & (\red{94.0}, \blue{92.9}) & (\blue{94.3}, \red{92.7}) & (93.4, 92.2) & (92.9, 91.9) \\
			& VisA & (78.1, 77.5) & (79.8, 84.3) & (82.1, 85.4) & (83.0, 84.9) & (\blue{85.6}, \blue{88.5}) & (\red{85.5}, \red{88.1}) & (83.0, 86.0) & (84.1, 86.7) \\
			\cmidrule{1-10}
			\multirow{3}{*}{\begin{tabular}{@{}c@{}}Medical \\ (Image-level)\end{tabular}} & BrainMRI & (82.0, 90.7) & (82.8, 85.5) & (86.1, 92.3) & (84.9, 94.2) & (\blue{89.1}, \red{97.2}) & (86.8, 96.9) & (\red{88.1}, \blue{97.3}) & (87.3, 97.1) \\
			& Liver CT & (64.2, 55.9) & (62.7, 51.6) & (61.6, 53.1) & (64.2, 56.7) & (60.2, 54.2) & (\red{65.8}, 55.3) & (64.4, \red{57.3}) & (\blue{68.4}, \blue{58.9}) \\
			& Retina OCT & (42.5, 50.9) & (67.9, 71.3) & (75.7, 77.4) & (\red{82.7}, 80.3) & (\blue{84.4}, \blue{85.6}) & (81.1, \red{80.9}) & (80.3, 79.1) & (82.0, 79.7) \\
			\bottomrule
		\end{tabular}
		\vspace{-2em}
	\end{table*}
	
	\vspace{-1em}
	\section{Experiments}
	\label{sec:experiments}
	\vspace{-0.5em}
	To demonstrate the efficacy and robustness of our ACD-CLIP framework, we conducted a comprehensive evaluation. This involved benchmarking our model against state-of-the-art (SOTA) competitors across a wide range of datasets and performing in-depth ablation studies to quantify the contribution of each individual component.
	
	\vspace{-1em}
	\subsection{Experimental Setup}
	\vspace{-0.5em}
	\noindent\textbf{Datasets and Baselines.} We perform a comprehensive evaluation on 11 public benchmarks spanning the industrial domain (MVTec-AD~\cite{bergmann2019mvtec}, VisA~\cite{zou2022spot}, BTAD~\cite{mishra2021vt}, MPDD~\cite{jezek2021deep}, RSDD~\cite{yu2018coarse}) and the medical domain (BrainMRI, Liver CT, and Retina OCT from BMAD~\cite{bao2024bmad}, ColonDB~\cite{tajbakhsh2015automated}, ClinicDB~\cite{bernal2015wm}, Kvasir~\cite{jha2019kvasir}). We benchmark against recent SOTA methods, including WinCLIP~\cite{winclip}, CLIP-AD~\cite{clipad}, AnomalyCLIP~\cite{zhou2023anomalyclip}, and AdaCLIP~\cite{adaclip}.
	
	\noindent\textbf{Implementation Details.} Our framework is built upon the pre-trained CLIP model with a ViT-L/14 backbone at a 336px resolution. To maintain a strict zero-shot setting, we train our model on an auxiliary dataset with no category overlap (VisA for all non-VisA benchmarks, and MVTec-AD for the VisA benchmark). We report the standard Area Under the Receiver Operating Characteristic curve (AUROC) and Average Precision (AP) for both pixel-level and image-level tasks.
	
	\noindent\textbf{Complexity and Efficiency.} ACD-CLIP is highly efficient, totaling 439M parameters, with the Conv-LoRA adapter adding only 9.96M (2.3\%) trainable parameters. At a 518px resolution, the model requires 517.98 GFLOPs and consumes 3.6GB of GPU memory. With an inference latency of 105.91ms on an RTX A6000, it is 2--3$\times$ faster than WinCLIP (357.8ms), AnomalyCLIP (214.73ms), and CLIP-AD (261.4ms), ensuring high practical deployment feasibility.

	\begin{table}[!t]
		\centering
		\small
		\setlength{\tabcolsep}{1em}
		\vspace{-0.75em}
		\caption{Ablation study of our proposed components on the MVTec-AD dataset with a two-group configuration ($N = 2$). We report the average AUROC (\%) for both pixel-level and image-level anomaly detection.}
		\label{tab:ablation}
		\begin{tabular}{l|c|c}
			\hline
			\multicolumn{1}{c|}{\multirow{2}{*}{\textbf{Configuration}}} & \multicolumn{2}{c}{\textbf{Avg. AUROC}} \\
			\cline{2-3}
			& \textbf{Pixel-Level} & \textbf{Image-Level} \\
			\hline
			1. Baseline & 82.3 & 81.2 \\
			2. + Conv-LoRA Adapter & 89.1 (+6.8) & 84.1 (+2.9) \\
			3. + DFG & 87.6 (+5.3) & 85.9 (+4.7) \\
			\hline
			\textbf{4. Ours (ACD-CLIP)} & \textbf{91.7} & \textbf{90.9} \\
			\hline
		\end{tabular}
		\vspace{-2em}
	\end{table}

	\vspace{-1em}
	\subsection{Main Results and Analysis}
	\vspace{-0.5em}
	As shown in Table~\ref{tab:main_results}, our ACD-CLIP framework consistently establishes a new state-of-the-art across both industrial and medical domains. On the widely-used MVTec-AD benchmark, our model excels at fine-grained localization, boosting the pixel-level AP by nearly 10 percentage points over AnomalyCLIP. This superiority in capturing precise anomaly boundaries is a direct result of our architectural co-design.
	
\begin{figure}[!t]
		\centering
		\includegraphics[width=0.8\linewidth]{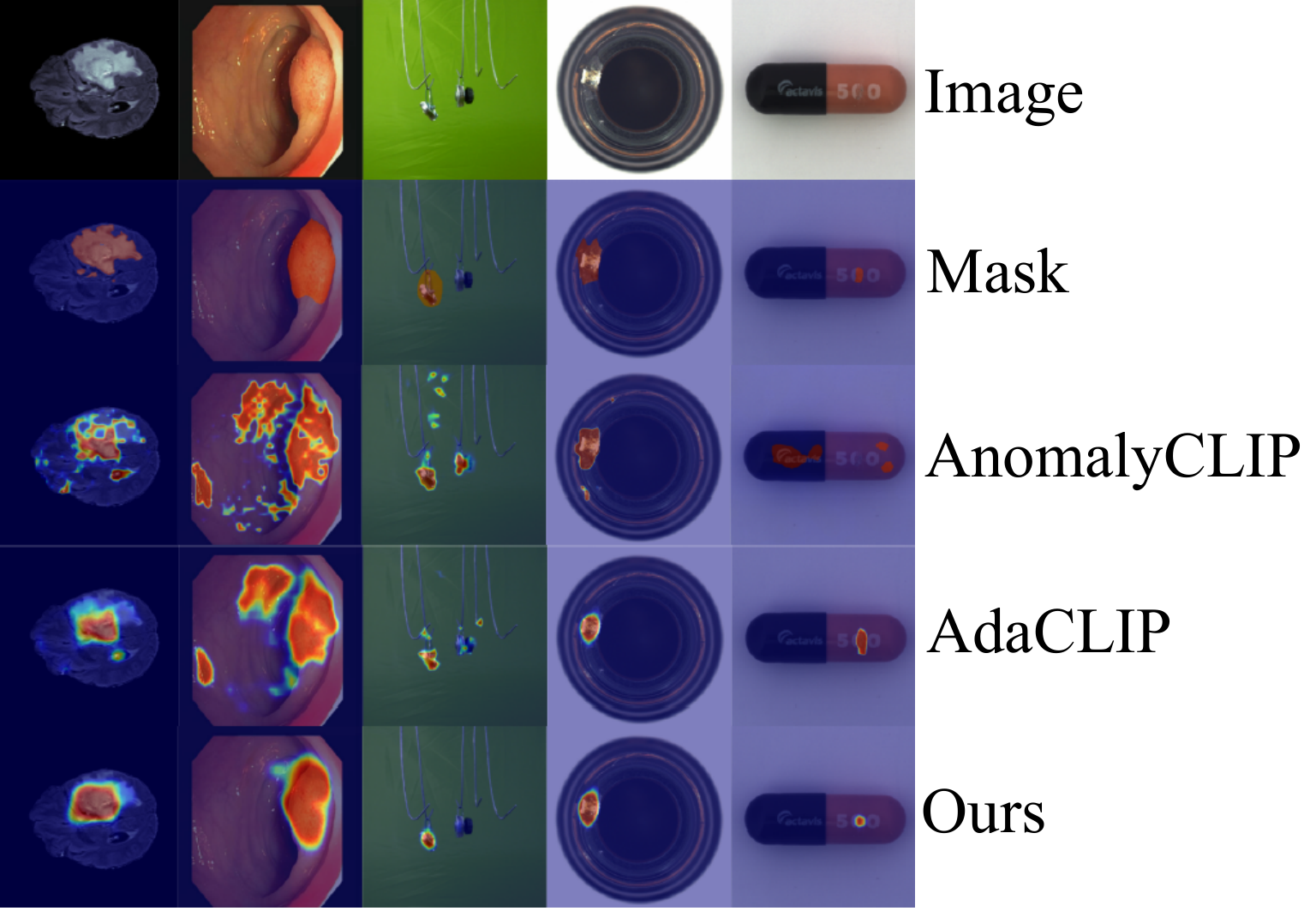}
		\vspace{-1em}
		\caption{Qualitative comparison on diverse industrial and medical datasets, showing our method's superior localization accuracy and noise suppression over state-of-the-art baselines.}
		\label{fig:3}
		\vspace{-1.5em}
	\end{figure}

	Trained only on industrial data, ACD-CLIP shows strong cross-domain generalization, achieving 90.4\% and 96.6\% pixel-level AUROC on ClinicDB and BrainMRI, respectively. Qualitative results (Figure~\ref{fig:3}) substantiate these gains, showcasing cleaner and more precise anomaly maps. We observe that the temperature $\tau$ in Eq.~\eqref{eq:anomaly_map} is key to this adaptability, with $\tau=0.05$ being optimal for sharp industrial boundaries and $\tau=0.075$ effectively capturing the diffused boundaries common in medical images.

	Results across groups ($N$) show performance peaks at $N=3$, balancing hierarchical representation specialization and complexity without overfitting on auxiliary data.

	\vspace{-1em}
	\subsection{Ablation Study}
	\vspace{-0.5em}
	Our ablation study (Table~\ref{tab:ablation}) validates each component's contribution against a baseline using standard LoRA and static fusion. Integrating the \textbf{Conv-LoRA Adapter} alone boosts the pixel-level AUROC by 6.8\%, confirming its critical role in providing the local inductive biases necessary for dense prediction. Separately, DFG improves pixel-level AUROC by 5.3\%; high cross-image variance in gating weights confirms sample-specific adaptive policies rather than static mapping. The full ACD-CLIP model achieves the highest performance, confirming the powerful synergy between our representation and fusion modules.
	
	\vspace{-1.5em}
	\section{CONCLUSION}
	\label{sec:conclusion}
	\vspace{-1em}
	
	We proposed ACD-CLIP, an Architectural Co-Design framework addressing VLM limitations in Zero-Shot Anomaly Detection. By synergizing a parameter-efficient \textbf{Conv-LoRA adapter} for local priors with a \textbf{Dynamic Fusion Gateway} for adaptive cross-modal interaction, our model extracts fine-grained, context-aware features. ACD-CLIP achieves state-of-the-art performance across diverse industrial and medical benchmarks. This work validates that joint representation-fusion adaptation is critical for dense perception tasks, offering a robust strategy that can extend to zero-shot segmentation and open-vocabulary detection.
	
	\vfill\pagebreak
	\label{sec:refs}
	{ 
		\small
		\begin{spacing}{0.85}
			\bibliographystyle{IEEEbib}
			\bibliography{inproceedings,article}

@article{bernal2015wm,
	title={WM-DOVA maps for accurate polyp highlighting in colonoscopy: Validation vs. saliency maps from physicians},
	author={Bernal, Jorge and S{\'a}nchez, F Javier and Fern{\'a}ndez-Esparrach, Gloria and Gil, Debora and Rodr{\'\i}guez, Cristina and Vilari{\~n}o, Fernando},
	journal={Computerized medical imaging and graphics},
	volume={43},
	pages={99--111},
	year={2015},
	publisher={Elsevier}
}

@article{yu2018coarse,
	title={A coarse-to-fine model for rail surface defect detection},
	author={Yu, Haomin and Li, Qingyong and Tan, Yunqiang and Gan, Jinrui and Wang, Jianzhu and Geng, Yangli-ao and Jia, Lei},
	journal={IEEE Transactions on Instrumentation and Measurement},
	volume={68},
	number={3},
	pages={656--666},
	year={2018},
	publisher={IEEE}
}

@article{tajbakhsh2015automated,
	title={Automated polyp detection in colonoscopy videos using shape and context information},
	author={Tajbakhsh, Nima and Gurudu, Suryakanth R and Liang, Jianming},
	journal={IEEE transactions on medical imaging},
	volume={35},
	number={2},
	pages={630--644},
	year={2015},
	publisher={IEEE}
}

@article{vit,
	title={An image is worth 16x16 words: Transformers for image recognition at scale},
	author={Dosovitskiy, Alexey and Beyer, Lucas and Kolesnikov, Alexander and Weissenborn, Dirk and Zhai, Xiaohua and Unterthiner, Thomas and Dehghani, Mostafa and Minderer, Matthias and Heigold, Georg and Gelly, Sylvain and others},
	journal={arXiv preprint arXiv:2010.11929},
	year={2020}
}

@article{lora,
	title={Lora: Low-rank adaptation of large language models.},
	author={Hu, Edward J and Shen, Yelong and Wallis, Phillip and Allen-Zhu, Zeyuan and Li, Yuanzhi and Wang, Shean and Wang, Lu and Chen, Weizhu and others},
	journal={ICLR},
	volume={1},
	number={2},
	pages={3},
	year={2022}
}

@article{flamingo,
	title={Flamingo: a visual language model for few-shot learning},
	author={Alayrac, Jean-Baptiste and Donahue, Jeff and Luc, Pauline and Miech, Antoine and Barr, Iain and Hasson, Yana and Lenc, Karel and Mensch, Arthur and Millican, Katherine and Reynolds, Malcolm and others},
	journal={Advances in neural information processing systems},
	volume={35},
	pages={23716--23736},
	year={2022}
}

@article{liu2023visual,
	title={Visual instruction tuning},
	author={Liu, Haotian and Li, Chunyuan and Wu, Qingyang and Lee, Yong Jae},
	journal={Advances in neural information processing systems},
	volume={36},
	pages={34892--34916},
	year={2023}
}

@article{zhou2023anomalyclip,
	title={Anomalyclip: Object-agnostic prompt learning for zero-shot anomaly detection},
	author={Zhou, Qihang and Pang, Guansong and Tian, Yu and He, Shibo and Chen, Jiming},
	journal={arXiv preprint arXiv:2310.18961},
	year={2023}
}

@article{convlora,
	title={Convolution meets lora: Parameter efficient finetuning for segment anything model},
	author={Zhong, Zihan and Tang, Zhiqiang and He, Tong and Fang, Haoyang and Yuan, Chun},
	journal={arXiv preprint arXiv:2401.17868},
	year={2024}
}

@inproceedings{long2015fully,
	title={Fully convolutional networks for semantic segmentation},
	author={Long, Jonathan and Shelhamer, Evan and Darrell, Trevor},
	booktitle={Proceedings of the IEEE conference on computer vision and pattern recognition},
	pages={3431--3440},
	year={2015}
}

@inproceedings{ronneberger2015u,
	title={U-net: Convolutional networks for biomedical image segmentation},
	author={Ronneberger, Olaf and Fischer, Philipp and Brox, Thomas},
	booktitle={International Conference on Medical image computing and computer-assisted intervention},
	pages={234--241},
	year={2015},
	organization={Springer}
}

@inproceedings{milletari2016v,
	title={V-net: Fully convolutional neural networks for volumetric medical image segmentation},
	author={Milletari, Fausto and Navab, Nassir and Ahmadi, Seyed-Ahmad},
	booktitle={2016 fourth international conference on 3D vision (3DV)},
	pages={565--571},
	year={2016},
	organization={Ieee}
}

@inproceedings{lin2017focal,
	title={Focal loss for dense object detection},
	author={Lin, Tsung-Yi and Goyal, Priya and Girshick, Ross and He, Kaiming and Doll{\'a}r, Piotr},
	booktitle={Proceedings of the IEEE international conference on computer vision},
	pages={2980--2988},
	year={2017}
}

@inproceedings{jha2019kvasir,
	title={Kvasir-seg: A segmented polyp dataset},
	author={Jha, Debesh and Smedsrud, Pia H and Riegler, Michael A and Halvorsen, P{\aa}l and De Lange, Thomas and Johansen, Dag and Johansen, H{\aa}vard D},
	booktitle={International conference on multimedia modeling},
	pages={451--462},
	year={2019},
	organization={Springer}
}

@inproceedings{houlsby2019parameter,
	title={Parameter-efficient transfer learning for NLP},
	author={Houlsby, Neil and Giurgiu, Andrei and Jastrzebski, Stanislaw and Morrone, Bruna and De Laroussilhe, Quentin and Gesmundo, Andrea and Attariyan, Mona and Gelly, Sylvain},
	booktitle={International conference on machine learning},
	pages={2790--2799},
	year={2019},
	organization={PMLR}
}

@inproceedings{bergmann2019mvtec,
	title={MVTec AD--A comprehensive real-world dataset for unsupervised anomaly detection},
	author={Bergmann, Paul and Fauser, Michael and Sattlegger, David and Steger, Carsten},
	booktitle={Proceedings of the IEEE/CVF conference on computer vision and pattern recognition},
	pages={9592--9600},
	year={2019}
}

@inproceedings{jezek2021deep,
	title={Deep learning-based defect detection of metal parts: evaluating current methods in complex conditions},
	author={Jezek, Stepan and Jonak, Martin and Burget, Radim and Dvorak, Pavel and Skotak, Milos},
	booktitle={2021 13th International congress on ultra modern telecommunications and control systems and workshops (ICUMT)},
	pages={66--71},
	year={2021},
	organization={IEEE}
}

@inproceedings{mishra2021vt,
	title={VT-ADL: A vision transformer network for image anomaly detection and localization},
	author={Mishra, Pankaj and Verk, Riccardo and Fornasier, Daniele and Piciarelli, Claudio and Foresti, Gian Luca},
	booktitle={2021 IEEE 30th International Symposium on Industrial Electronics (ISIE)},
	pages={01--06},
	year={2021},
	organization={IEEE}
}

@inproceedings{clip,
	title={Learning transferable visual models from natural language supervision},
	author={Radford, Alec and Kim, Jong Wook and Hallacy, Chris and Ramesh, Aditya and Goh, Gabriel and Agarwal, Sandhini and Sastry, Girish and Askell, Amanda and Mishkin, Pamela and Clark, Jack and others},
	booktitle={International conference on machine learning},
	pages={8748--8763},
	year={2021},
	organization={PmLR}
}

@inproceedings{defard2021padim,
	title={Padim: a patch distribution modeling framework for anomaly detection and localization},
	author={Defard, Thomas and Setkov, Aleksandr and Loesch, Angelique and Audigier, Romaric},
	booktitle={International conference on pattern recognition},
	pages={475--489},
	year={2021},
	organization={Springer}
}

@inproceedings{zou2022spot,
	title={Spot-the-difference self-supervised pre-training for anomaly detection and segmentation},
	author={Zou, Yang and Jeong, Jongheon and Pemula, Latha and Zhang, Dongqing and Dabeer, Onkar},
	booktitle={European conference on computer vision},
	pages={392--408},
	year={2022},
	organization={Springer}
}

@inproceedings{roth2022towards,
	title={Towards total recall in industrial anomaly detection},
	author={Roth, Karsten and Pemula, Latha and Zepeda, Joaquin and Sch{\"o}lkopf, Bernhard and Brox, Thomas and Gehler, Peter},
	booktitle={Proceedings of the IEEE/CVF conference on computer vision and pattern recognition},
	pages={14318--14328},
	year={2022}
}

@inproceedings{winclip,
	title={Winclip: Zero-/few-shot anomaly classification and segmentation},
	author={Jeong, Jongheon and Zou, Yang and Kim, Taewan and Zhang, Dongqing and Ravichandran, Avinash and Dabeer, Onkar},
	booktitle={Proceedings of the IEEE/CVF Conference on Computer Vision and Pattern Recognition},
	pages={19606--19616},
	year={2023}
}

@inproceedings{adaclip,
	title={Adaclip: Adapting clip with hybrid learnable prompts for zero-shot anomaly detection},
	author={Cao, Yunkang and Zhang, Jiangning and Frittoli, Luca and Cheng, Yuqi and Shen, Weiming and Boracchi, Giacomo},
	booktitle={European Conference on Computer Vision},
	pages={55--72},
	year={2024},
	organization={Springer}
}

@inproceedings{clipad,
	title={Clip-ad: A language-guided staged dual-path model for zero-shot anomaly detection},
	author={Chen, Xuhai and Zhang, Jiangning and Tian, Guanzhong and He, Haoyang and Zhang, Wuhao and Wang, Yabiao and Wang, Chengjie and Liu, Yong},
	booktitle={International Joint Conference on Artificial Intelligence},
	pages={17--33},
	year={2024},
	organization={Springer}
}

@incollection{jie2024convolutional,
	title={Convolutional bypasses are better vision transformer adapters},
	author={Jie, Shibo and Deng, Zhi-Hong and Chen, Shixuan and Jin, Zhijuan},
	booktitle={ECAI 2024},
	pages={202--209},
	year={2024},
	publisher={IOS Press}
}

@inproceedings{bao2024bmad,
	title={Bmad: Benchmarks for medical anomaly detection},
	author={Bao, Jinan and Sun, Hanshi and Deng, Hanqiu and He, Yinsheng and Zhang, Zhaoxiang and Li, Xingyu},
	booktitle={Proceedings of the IEEE/CVF Conference on Computer Vision and Pattern Recognition},
	pages={4042--4053},
	year={2024}
}
		\end{spacing}
	}
	
\end{document}